\documentclass[10pt,twocolumn,letterpaper]{article}

\pdfoutput=1
\usepackage{iccv}
\usepackage{times}
\usepackage{epsfig}
\usepackage{graphicx}
\usepackage{amsmath}
\usepackage{amssymb}
\usepackage{diagbox}
\usepackage{booktabs}
\usepackage{tabularx,booktabs,caption}
\usepackage{multirow}
\usepackage{stfloats}
\usepackage{subcaption}
\usepackage{footnote}

\usepackage[pagebackref=true,breaklinks=true,letterpaper=true,colorlinks,bookmarks=false]{hyperref}

\iccvfinalcopy 

\def\iccvPaperID{2093} 

\begin{document}

\title{Face Age Progression With Attribute Manipulation}

\author{Sinzith Tatikonda\\
{\tt\small sinzithtatikonda@gmail.com}
\and
Athira Nambiar\\
{\tt\small anambiar@cse.iitm.ac.in}
\and
Anurag Mittal\\
{\tt\small amittal@cse.iitm.ac.in}
\and
\small Department of Computer Science and Engineering, \hspace{1mm}\small Indian Institute of Technology Madras, India\\

}

\maketitle
\ificcvfinal\thispagestyle{empty}\fi

\begin{abstract}
Face is one of the predominant means of person recognition. In the process of ageing, human face is prone to many factors such as time, attributes, weather and other subject specific variations. The impact of these factors were not well studied in the literature of face aging. In this paper, we propose a novel holistic model in this regard \textit{viz.,} ``Face Age progression With Attribute Manipulation (FAWAM)", i.e. generating face images at different ages while simultaneously varying attributes and other subject specific characteristics. We address the task in a bottom-up manner, as two submodules i.e. face age progression and face attribute manipulation. For face aging, we use an attribute-conscious face aging model with a pyramidal generative adversarial network that can model age-specific facial changes while maintaining intrinsic subject specific characteristics. For facial attribute manipulation, the age processed facial image is manipulated with desired attributes while preserving other details unchanged, leveraging an attribute generative adversarial network architecture. We conduct extensive analysis in standard large scale datasets and our model achieves significant performance both quantitatively and qualitatively. 
\end{abstract}

\section{Introduction}
In this paper, we address a novel \textit{``Face Age Progression With Attribute Manipulation"} task i.e., upon a given young face input image, aesthetically render the effects of facial aging along with simultaneously manipulating facial attributes. This has great practical application in forensics, entertainment, investigation etc. With the development of deep convolutional neural networks (CNNs) and large scale labeled datasets, significant advances have been made recently in Computer Vision. Albeit facial aging and attribute manipulation has been addressed independently as two active research areas in Vision, their impact on each other hasn't studied very well.
This work investigates some initial comprehensive study in this direction.

\begin{figure}[t]
\small 
\captionsetup{size=small}
\begin{center}
   \includegraphics[width=0.8\linewidth]{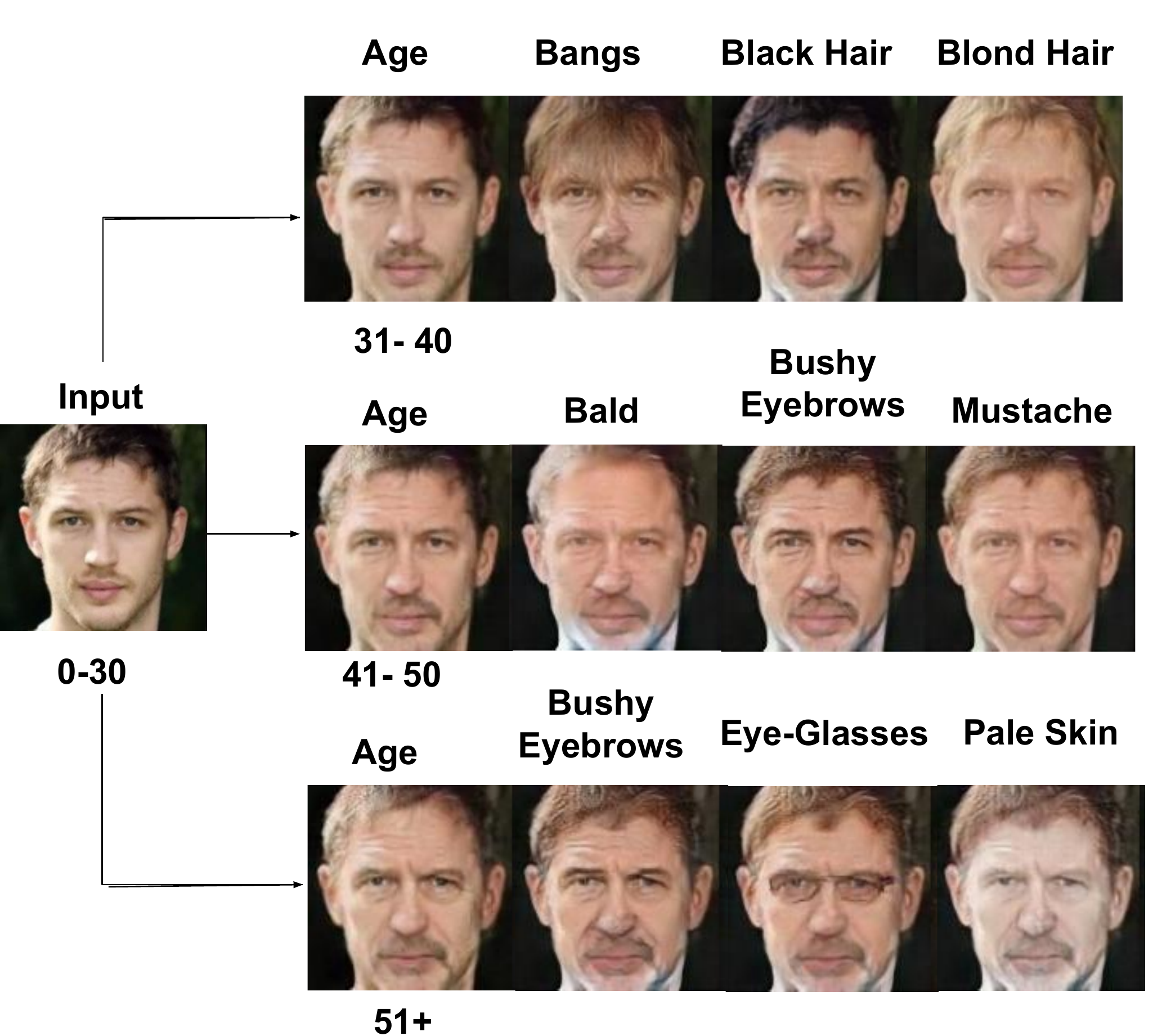}
\end{center}
\vspace{-0.5cm}
   \caption{Demonstration of our ``Face aging with attribute manipulation" (FAWAM) results. Images in each row are the attribute manipulated output images corresponding to different age clusters.}
\vspace{-0.5cm}
\label{fig:demo}
\end{figure}

The main motivation of this work is from basic human recognition ability \cite{wilmer2014face}, as humans can recognize their peers even if they meet after a long period of time. 
To understand how does the machine recognize this, it was required to understand the process of aging and facial attribute analysis. Regarding the face aging problem, many research were going on in the recent past from identifying the age of person from a facial image \cite{facepp} to producing different age images of single person \cite{liu2019attribute}. There has been a rapid growth in facial attribute analysis too e.g., manipulating facial attributes \cite{shen2017learning}, producing animated filters on faces \cite{zhang2018generative}, etc..

The intrinsic complexity of physical aging, the influences from other external factors such as pose, illumination, and expression (PIE) variations and the shortage of labeled aging data collectively make face age progression a difficult problem. Two important key targets of face aging are to achieve aging accuracy and identity permanence \cite{yang2018learning}. After the early attempts based on skin’s anatomical structure and mechanical simulations, data-driven approaches gained more popularity  either by applying the prototype of aging details to test faces \cite{kemelmacher2014illumination}\cite{suo2009compositional} or by modeling the relationship between facial changes and corresponding ages \cite{suo2012concatenational}\cite{park2010age}\cite{wang2012combining}. Further, significant improvement on the aging results were reported while using
deep generative networks towards image generation \cite{dosovitskiy2016generating}\cite{goodfellow2014generative}\cite{isola2017image}
with more appealing aging effects and less ghosting artifacts. Nevertheless, since such approaches model face transformation across two age groups, age factor gains prominent role than the identity information, thus curbing both the goals to achieve simultaneously.
Further, they require multiple face images of different ages of the same person at the training stage thus leading to another practical challenge of intra-individual aging dataset collection.

The facial attribute manipulation problem addresses face editing via manipulating single or multiple attributes such as hair color, glasses, expression, mustache etc. One major challenge associated with this task is the difficulty in collecting annotated images of a particular person with varying attributes. However, with the arrival of deep neural network techniques and large scale datasets\cite{liu2015faceattributes}, considerable advancements in the field has been witnessed. In particular, generative models such as variational auto-encoder (VAE) and generative adversarial network (GAN) are employed to bring out significant facial attribute editing. The encoder-decoder architecture is one of the most effective solutions for using a single model for multiple attribute manipulation. To this end, many key works such as VAE/GAN \cite{larsen2016autoencoding}, IcGAN\cite{perarnau2016invertible}, Fader networks\cite{lample2017fader}, AttGAN\cite{he2019attgan} etc. were reported in the past. Nonetheless, such works
considered age only as an attribute element and consequently, age gains only very little importance compared to the attribute factor.

In this work, we address a high-level and generic task named ``Face age progression with attribute manipulation" wherein both the aging and attribute manipulation counterparts get equal roles to play, with which the model can exclusively render a particular attribute for a specific age group. To this end, we propose a novel comprehensive architecture viz.,  ``Face aging with attribute manipulation" (FAWAM) model by combining both the Face Aging (FA) module and the Attribute manipulation (AM) module in a sequential manner. In particular, we implement FA module by using pyramidal GAN architecture \cite{yang2018learning}. This method works on different face samples having different variations in posture, expression, etc., and real life-like aging effect results are achieved. Regarding Face attribute editing component, we leverage AttGAN\cite{he2019attgan} to incorporate the attribute editing upon the face image. We perform editing by using binary facial attributes and apply the attribute classification constraint to the generated image to guarantee the correct change of desired attributes. The reconstruction learning is introduced to preserve attribute-excluding details and the adversarial learning is employed for photorealistic editing.

Extensive baseline studies, performance analysis and ablation studies are carried out on CelebA, CACD, FGNET and UTK-face datasets. During training, each module is trained independently on its respective dataset, whereas while testing, end-to-end model inference is performed. 

The key contributions of this study include:
(1) Proposal of a novel ``Face aging with attribute manipulation”(FAWAM) model that bestows signficant roles for age and attribute components during image synthesis; a task not well explored in the past.
(2) Extensive expeiments on multiple datasets that verify the effectiveness and robustness of our FAWAM model: Especially, the influence of age on attribute clearly enhances the synthesised image quality.

\section{Related works}
\subsection{Face aging:}
For a particular person, some facial properties tend to be constant over time, and some age specific features tend to change over time.
Hence, it is required to differentiate and predict all those changes through the face aging model. According to recent studies, aging in humans can be split into two stages, child and adult aging \cite{antipov2016apparent}. We focus our work only in adult faces, since the former is prone to various natural modifications in the initial years. 

The early works on aging exploited physical models and skin anatomy, 
to simulate the aging progression of facial muscles. For instance, Todd et. al. \cite{todd1980perception} proposed head growth modelling via a computable geometric procedure and Wu et al. \cite{wu1994plastic} proposed a multi-layered dynamic skin model to simulate wrinkles. Later, many data-driven approaches came to prominence where they leveraged training faces to learn aging pattern, rather than relying much on the biological information. In this regard, many recent works exploited GANs to improve the results by synthesising faces with corresponding aging factors. Wang et al. \cite{wang2012combining} proposed a model by mapping the faces in a tensor space and the age-related details were added on top of that. Similarly, Yang et al. \cite{yang2016face} tackled it as a multi-attribute decomposition problem, and the age progression was incorporated by manipulating the age component to the target age group.
Further, deep generative networks were used for generating smooth versions of real-looking faces across different ages, for e.g. \cite{wang2016recurrent} and S2GAN\cite{he2019s2gan}. Temporal Non-Volume Preserving aging approach \cite{nhan2017temporal} realised the short-term age progression
by mapping the data densities of two consecutive age groups with ResNet blocks and the long-term aging was finally achieved by combining short-term stages. However, the limitation of such approaches was the variation in colour/ identity/ expression etc., since it is mainly dependent on the probability distribution rather than individual information. One of the most popular recent work leveraged pyramidal architecture of GANs \cite{yang2018learning} towards age transformation, also by taking the advantage from individual-dependent critic to keep the identity cue stable, as well as multi-pathway discriminator to refine aging details. We use Pyramidal GAN architecture in the face aging submodule. 

\begin{figure*}[!htb]\centering
\small 
\captionsetup{size=small}

   \begin{minipage}{1\textwidth}
     \includegraphics[width=\linewidth]{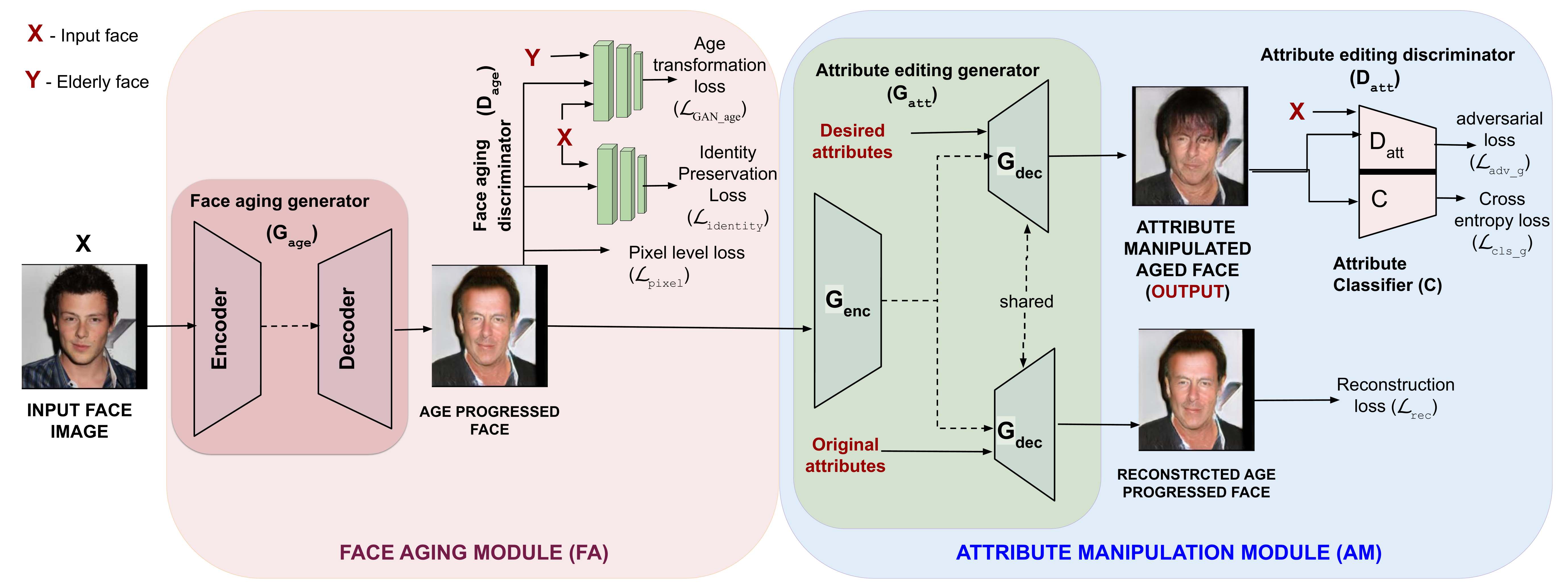}
     \caption{ Framework of the proposed Face Age Progression With Attribute Manipulation (FAWAM) model. It has face aging module and attribute manipulation module. The inner architectures of both modules comprises of GAN architectures. }
     \label{arch}
   \end{minipage}  
\end{figure*}

\subsection{Facial Attribute Manipulation:}
Editing/ manipulation of facial attributes is achieved via either of the following methods: (a) Optimization based methods\cite{upchurch2017deep,li2016convolutional} or (b) Learning based methods\cite{li2016deep,lample2017fader,choi2018stargan}. The former optimization method uses CNN based architectures and they try to minimize the attribute loss of given face and set of different attribute faces as in CNAI \cite{li2016convolutional}. Similarly, DFI \cite{upchurch2017deep} moves the deep feature of input face linearly along the direction between the faces with and without the expected attributes and the input face is optimized to match its deep feature with the moved feature. However, the training of such optimization-based methods is laborious and time consuming. 
The latter approach i.e. Learning based methods employ residual learning strategy that minimizes the objective loss function. These methods are popular than the optimization methods. In \cite{li2016deep}, Li et. al. proposed a model to edit attribute in a face image using an adversarial attribute loss and a deep identity feature loss. Another work \cite{shen2017learning} leverages dual residual learning strategy to train two networks, simultaneously to add/remove a specific attribute. One downside of such learning methods is the necessity of different models for different attributes(or combinations). To overcome this practical constraint, single model for multiple facial attribute editing was proposed. For e.g., attribute manipulation is achieved in VAE/GAN \cite{larsen2016autoencoding} by modifying the latent representation. In Fader Networks \cite{lample2017fader}, an adversarial process upon the latent representation of an autoencoder is carried out to learn the attribute-invariant representation. Further, the decoder generates the edited result based on the arbitrary attribute vector. Another work viz., AttGAN \cite{he2019attgan} uses an encoder-decoder architecture that is inspired by the VAE/GAN \cite{larsen2016autoencoding} and Fader Networks \cite{lample2017fader}. It models the relation between the latent representation and the attributes. In this work, we use AttGAN\cite{he2019attgan} as the facial attribute editing module.
Some other works in this area are StarGAN \cite{choi2018stargan}, STGAN\cite{liu2019stgan}, SSCGAN\cite{chu2020sscgan} and SG-GAN\cite{zhang2018sparsely}.

\section{Proposed architecture for Face Age Progression With Attribute Manipulation: }

In this section, we describe our novel architecture designed towards the task of ``Face age progression with attribute manipulation" (FAWAM). In this pilot study, we try to tackle the problem in a sequential manner by fusing the individual components i.e. Face Aging (FA) module and Attribute Manipulation (AM) module. The holistic architecture of our propsed FAWAM model is explained in Sec. \ref{overview}. The internal architecture of  FA and AM submodules are described in detail in Sec \ref{sec:FA} and \ref{sec:FAM} respectively. 
\subsection{Overview of our proposed FAWAM model}\label{overview}
The key objective of our model is to generate the aged attribute manipulated face image as the output, while providing a young face and the desired attributes as the input. To achieve this goal, we address this complex task in a bottom-up fashion, by addressing it as two separate modules i.e. Face Aging (FA) and Attribute Manipulation (AM). Although we can treat \textit{age} as an attribute (viz., gender, glasses etc.), in this study we wish to provide equal importance (probability) to both the \textit{ageing} as well as \textit{attribute} factors. Hence, we treat them as separate modules and combine them in a comprehensive manner.
The overall architecture is shown in Fig. \ref{arch}. The base network for both FA and AM modules is Generative Adversarial Network (GAN) architecture. Initially, the model is fed with the young facial input image. It is given to the face aging module, which is internally comprised of face aging generator and a face aging discriminator. Using three key loss functions -i.e., Identity preservation loss for maintaining the identity, GAN loss for aging and pixel level loss in image space- the network is trained to produce the age progressed face as the output of FA module. Further, this aged image is given as the input to the attribute manipulation module. Here, in addition to the image, we also provide the attribute as the second input. Using both of these inputs, AM module, that internally consists of an attribute editing generator and attribute editing discriminator, produces the attribute manipulated aged face as the final output. To obtain these manipulations we employ adversarial loss, cross entropy loss and reconstruction loss. Though the modules FA and AM are trained independently on their respective datasets, for testing it is end-to-end model. More details on the internal architecture of each sub-module and the methodology adopted are described in detail in the sections \ref{sec:FA}, \ref{sec:FAM}.

\subsection{Face aging module:}\label{sec:FA}
We have implemented the Face Aging (FA) module by using pyramidal GAN architecture proposed by Yang et al.\cite{yang2018learning}. This model has CNN based architecture for generator ($G_{age}$) to learn the age transformation effects of face image, and the training critic has squared euclidean loss, GAN loss and identity loss. In discriminator, we have VGG-16\cite{simonyan2014very} structure which captures the properties from exact pixel values to high-level age-specific information, this module leverages intrinsic pyramid hierarchy. The pyramid discriminator at multiple levels tries to advance and simulate the face aging effects. This method works on different face samples having different variations in posture, expression, etc..\\

\noindent \textbf{Generator ($G_{age}$):}\\
The Generator takes young face images as input, and generates elderly(age-progressed) face images as output. The generator is a combination of encoder and decoder with residual blocks in-between as shown in Fig. \ref{arch}. The input image is first passed through three convolutional layers for encoding it into latent space. This would capture the facial properties that tends to be constant with time. Then we have four residual blocks for capturing similar features shared by input and output faces. Then a decoder having three deconvolutional layers for the age progression to the target image space. We have got $3$x$3$ convolutional kernels with a stride of two, to confirm that each pixel contributes and neighbouring pixels would be modified in a synergistic manner. Paddings are added to make sure that input and output have equivalent size. Every convolutional layers is followed by Instance Normalization and ReLU non-linearity activation.

\noindent \textbf{Discriminator ($D_{age}$):}\\
The input to discriminator ( $D_{age}$) would be young faces, elderly faces and output of Generator (generated images). The discriminator is termed as system critic because it has previous knowledge of the data density. The Discriminator network $D_{age}$ is so introduced, such that it outputs a scalar quantity $D_{age}(x)$ representing with what probability $x$ comes from the data. The distribution of the generated faces $P_g$ (we would denote the distribution of young face images as $x\sim P_{young}$, then generated ones as $G_{age}(x)\sim P_g$) is meant to be equivalent to the distribution $P_{old}$ once optimality is reached. To attain more satisfying and realistic age specific facial details, the true young face images and the generated age transformed  face images  are  passed to $D_{age}$ as negative samples and the real elderly face images as positive samples. The training process will minimize the below loss functions alternately.

$\phi_{age}$ is pre-trained for multi label classification task of age estimation for the VGG-16 structure \cite{simonyan2014very} and after convergence, we remove the fully connected layers and integrate it into the framework. The detailed architecture of FA module can be found in the \textbf{supplementary material}.\\

\noindent \textbf{Objective function}\\
The objective function of face aging module is combination of pixel level loss ($\mathcal{L}_{pixel}$), identity preservation loss ($\mathcal{L}_{identity}$), and Age transformation loss ($\mathcal{L}_{GAN_{G_{age}}}$)  i.e. discriminator loss. 
\begin{equation}
    \mathcal{L}_{G_{age}}=\lambda_a \mathcal{L}_{GAN_{G_{age}}}+\lambda_p\mathcal{L}_{pixel}+
\lambda_i\mathcal{L}_{identity}
\end{equation}
\begin{equation}
    \mathcal{L}_{D_{age}}=\mathcal{L}_{GAN_{D_{age}}}
\end{equation}

We train $G_{age}$ and $D_{age}$ iteratively till they will learn age transformation effects and then finally $G_{age}$ would learn the transformation and $D_{age}$ would become a reliable reckoner.

\subsection{Attribute manipulation module: }\label{sec:FAM}
Face attributes are interesting due to their detailed description of human faces. Manipulating these attributes would help us to think of how the person looks at different stages. Even if we manipulate the attributes, the generated images should retain most of the details in attribute-irrelevant areas. This area has a major utility in entertainment industry.

Face attribute editing is the task that tries to manipulates face image for given attribute value. The desired result is that only attribute determined region should be manipulated while the remaining portion of image remains unmodified. We are using generative adversarial network (GAN) and encoder-decoder architecture for this task. By using the encoder-decoder architecture, face attribute editing is achieved by decoding the latent representation of the given face conditioned on the desired attributes. In this work, we apply attribute classification constraint to the generated image to verify the desired change of attributes. Then reconstruction learning is used to preserve attribute-excluding details. Besides, the adversarial learning is used for realistic changes. These three components cooperate with each other for facial attribute editing.\\

\noindent \textbf{Framework:}
The framework of face attribute editing module is shown in Fig. \ref{arch}. It has GAN architecture. The generator comprises of two basic sub-networks, i.e., an encoder $G_{att-enc}$ and a decoder $G_{att-dec}$. The discriminator($D_{att}$) consists a stack of convolutional layers followed by fully connected layers, and the classifier(C) has a similar architecture and shares all convolutional layers with $D_{att}$. The encoder($G_{att-enc}$) is a stack of convolutional layers and the decoder($G_{att-dec}$) is a stack of de-convolutional layers. The detailed architecture of AM module can be found in the \textbf{supplementary material}.\\ 

\noindent \textbf{Objective function}
The objective function of face attribute editing module is combination of attribute classification constraint($\mathcal{L}_{cls_g}$), the reconstruction loss($\mathcal{L}_{rec}$) and the adversarial loss($\mathcal{L}_{adv_g}$).

The objective for the encoder and decoder is formulated as below.
\begin{equation}
   \min_{G_{att-enc},G_{att-dec}}\mathcal{L}_{enc,dec}=\lambda_1 \mathcal{L}_{rec}+\lambda_2\mathcal{L}_{cls_g}+\mathcal{L}_{adv_g}
\end{equation}
and the objective for the discriminator and the attribute classifier is formulated as below,
\begin{equation}
    \min_{D_{att},C}\mathcal{L}_{dis,cls}=\lambda_3\mathcal{L}_{cls_c}+\mathcal{L}_{adv_d}
\end{equation}

\section{Experimental Results:}

\subsection{Datasets}
\subsubsection{Face aging:}
\noindent \textbf{CACD:} This dataset is proposed by \cite{chen2015face}. It contains 163,446 face images of 2,000 celebrities captured in less controlled conditions. Besides large variations in PIE, images in CACD are accumulated through Google Image Search, making it a difficult and challenging dataset. \\
\noindent \textbf{FGNET:} This dataset is proposed by \cite{fgnet}. It contains 1,002 images from 82 individuals. The age span is 0-69 years, there are 6-18 images per subject and average age is 15.54 years. It is used for testing. \\
\noindent \textbf{UTK-Face:} This dataset is proposed by \cite{zhifei2017cvpr}, it has 20,000 images of age ranging from 0 to 116 years. These images are collected from Google and Bing search engines. 68 landmarks are defined  with cropped and aligned faces. It is used for testing.
\subsubsection{Attribute Manipulation:}
\noindent \textbf{CelebA:}  CelebFaces Attributes Dataset (CelebA) is a large-scale face attributes dataset, proposed by \cite{liu2015faceattributes}, contains 202,599 number of face images of various celebrities, 10,177 unique identities, 40 binary attribute annotations per image and 5 landmark locations. Officially, CelebA is separated into training set, validation set and testing set. We use the training set and validation set together to train our model while using the testing set for evaluation. 

Thirteen attributes with strong visual impact are chosen in all our experiments, including “Bald”, “Bangs”, “Black Hair”, “Blond Hair”, “Brown Hair”, “Bushy Eyebrows”, “Eye-glasses”, “Gender”, “Mouth Open”, “Mustache”, “No Beard”, “Pale Skin”.

\subsection{Implementation details}
Regarding the FA module, We follow age span of 10 years  \cite{yang2016face} for every aging cluster, Therefore will train differently for different age clusters. The optimizer we use is Adam for both generator and discriminator, with the learning rate of $1$x$10^{-4}$, batch size of 8 and weight decay of $0.5$ for $50,000$ iterations. The trade-off parameters are set to $\lambda_p=0.20, \lambda_a=750.00$ and $\lambda_i=0.005$, for CACD. We update discriminator for each iteration, use the aging and identity critics for every generator iteration, and use the pixel level critic for each 5 generator iterations.

For the AM module, the model is trained by Adam optimizer \cite{kingma2014adam} $(\beta_1= 0.5, \beta_2= 0.999)$ with the batch size of 32 and the learning rate of 0.0002. The coefficients for the losses are set as: $\lambda_1= 100, \lambda_2= 10$, and $\lambda_3= 1$, which aims to make the loss values be in the same order of magnitude.

Regarding the training of our FAWAM model, both the FA and the AM modules are trained independently, whereas the testing of the holistic model is carried out in an end-to-end manner.

\renewcommand*{\thesubsubsection}{\roman{subsubsection}.}

\subsection{Face Aging module results:}
In this section, we describe the experimental analysis for the baseline Face Aging module. In particular, we carry out the aging performance evaluation of the FA module, both quantitatively as well as qualitatively. 
\vspace{-0.2cm}
\subsubsection{\textbf{Qualitative results:}} 
\begin{figure}[bt]
    \centering
    \small 
\captionsetup{size=small}

     \includegraphics[scale=0.4]{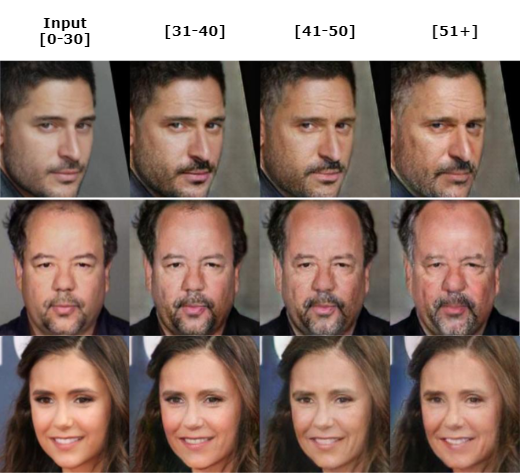}
     \caption{Aging effects on CACD Dataset for the baseline FA module. The first image in each row is the original face image and the next 3 images are in [31- 40], [41-50] and [51+] age clusters.}
     \label{Fig:FA_result}
\end{figure}

The visualizations of age progressed results on CACD are shown in Fig. \ref{Fig:FA_result}. We depict the result of face aging in different age clusters i.e., [31-40], [41-50], [51+] following most of the prior face aging analysis criteria in the field as in \cite{yang2018learning}\cite{liu2019attribute}. Results show that even with different facial attributes the model tries to give visually realistic and smooth aging effects.  
\vspace{-0.5cm}
\subsubsection{\textbf{Quantitative results:}}
We carry out two kinds of quantitative analysis for verification; i.e. the aging accuracy and the identity permanence. The former deals with vivid aging effects and their span, whereas the latter measures identity/familiarity between two face images. In order to measure these results, we use online face analysis tool FACE++ api\cite{facepp}. 

\begin{figure*}[bp]
    \centering
    \small 
\captionsetup{size=small}

   \begin{minipage}{\textwidth}
     \includegraphics[width=\linewidth]{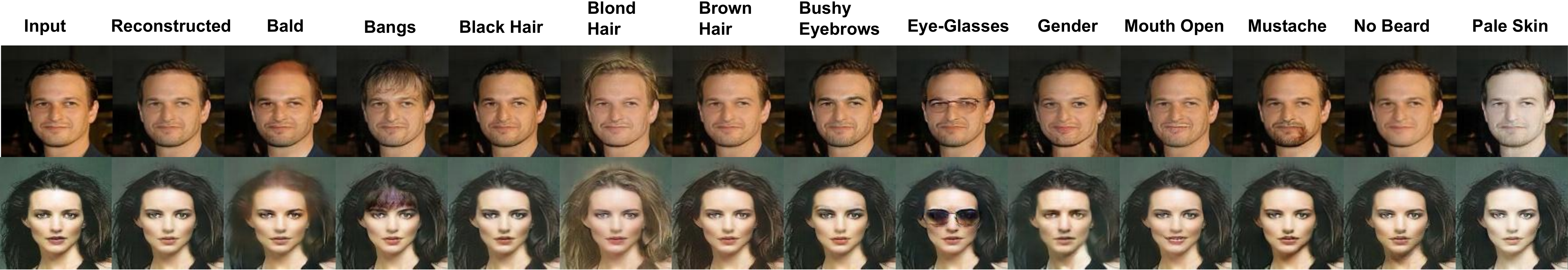}
     \caption{Results of facial attribute editing on CelebA dataset for the baseline AM module. For each attribute, editing is to invert the attribute, e.g., edit No-Beard to Beard etc.}
     \label{fam_result}
   \end{minipage}  
\end{figure*}

Identity preservation can be verified by using a pair of image that is input image and generated image, The result is verification confidence (probability value) between two images, i.e. how close the images are, higher the confidence higher the closeness to image. The result we expect is that value should not be close to 100 and rather not far, The values between 80 to 95 are considered good for different age ranges. The verification results are  given in the Table. \ref{table:FA_Verf} on CACD, FGNET and UTK face datasets. As can be seen in Table. \ref{table:FA_Verf}, as age increases the confidence decreases due to physical constraints occurred as part of aging.

\begin{table}[t]
    \centering
\small 
\captionsetup{size=small}
\caption{Face verification results for facial aging module (values are in \%)}
\label{table:FA_Verf}
    \begin{tabular}{c l c c c}
    \toprule
    \multirow{3}{*}{\bfseries Dataset} & 
    \multirow{3}{*}{\bfseries Test} & 
    \multicolumn{3}{c}{\bfseries Verification Confidence}\\ \cmidrule(lr){3-5}
    && Age[31-40] & Age[41-50] & Age[51+] \\ 
    \cmidrule(lr){1-5}
         &  Age[0-30]   &  92.1& 89.32 & 84.8\\
    CACD &  Age[31-40]  & - & 92.4 & 91.13 \\
         &  Age[41-50]  & - & - & 93.5 \\
    \cmidrule(lr){1-5}
         &  Age[0-30]   &   93.2 &  89.8 &  85.4 \\
    FGNET&  Age[31-40]  &   - & 91.3 & 89.2  \\
         &  Age[41-50]  &   - & - &  92.7  \\
    \cmidrule(lr){1-5}
    UTK &  Age[0-30]   &   91.2 &  89.8 &  85.4 \\
    \bottomrule
\end{tabular}
\end{table}

\begin{table}[t]
    \centering
\small 
\captionsetup{size=small}
\caption{Age estimation results for facial aging module (values are mean of predicted age).} 
\label{table:FA_age}
    \begin{tabular}{c c c c}
    \toprule
    {\bfseries Dataset} & 
    \multicolumn{3}{c}{\bfseries Age-Clusters}\\ 
    \cmidrule(lr){2-4}
    & [31-40] & [41-50] & [51+] \\ 
    \midrule
    CACD &   44.8$\pm$ 6.2 & 49.2$\pm$ 7.05 & 53.42$\pm$ 7.42 \\
    FGNET&     43.2$\pm$ 5.4 & 51.4$\pm$ 8.41 & 58.1$\pm$ 7.25  \\
    \bottomrule
\end{tabular}
\end{table}

The change in facial age should be in correspondence with real aging and the predicted result should be satisfying with the age cluster of it. We use face analysis tool Face++\cite{facepp} to verify aging accuracy. Age is calculated on each generated image from the respective age cluster and average is presented as accuracy. The results of aging accuracy are given in Table. \ref{table:FA_age} for CACD, FGNET datasets on the age clusters [31-40], [41-50], [51+]. We can observe that the predicted age clusters belongs to the intended category. We ascribe this may be due to the fact that the old-age related features such as wrinkles, hair colour etc. clearly place them in the correct category with higher precision (thanks to our pyramidal GAN for the realistic super smooth image rendering). However, in the 31+ category, the average age is slightly greater than the cluster's age, this is due to the data density of faces in their ages.The standard deviation is also calculated on all the synthesized faces.

\begin{figure}[bt]
\small 
\captionsetup{size=small}

\begin{center}
  \includegraphics[width=0.9\linewidth]{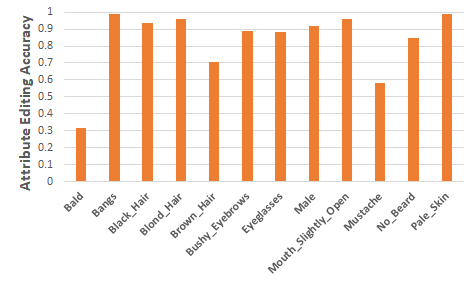}
\end{center}
\vspace{-0.9cm}
  \caption{Attribute editing accuracy of Attribute Manipulation module (Higher the Better). }
\label{fig:fam_acc}
\end{figure}

\subsection{Attribute Manipulation module results:}
In this section, we describe the experimental analysis for the baseline AM module. We conduct the attribute editing evaluation of the AM module in detail. 
\vspace{-0.2cm}
\subsubsection{\textbf{Qualitative results:}} Fig. \ref{fam_result} shows the result of facial attribute editing. These are for single attribute editing for each image. This model preserves all attribute excluding details like face identity, background and illumination. Our model accurately edits local and global attributes as can be seen in Fig. \ref{fam_result}. We have used 12 attributes for training on CelebA dataset. The attributes like eye-glasses, black hair, bangs, blond hair are edited more realistically.

\begin{figure*}[bp]
\small 
\captionsetup{size=small}

\begin{subfigure}{\linewidth}
\centering
  \includegraphics[width=\linewidth]{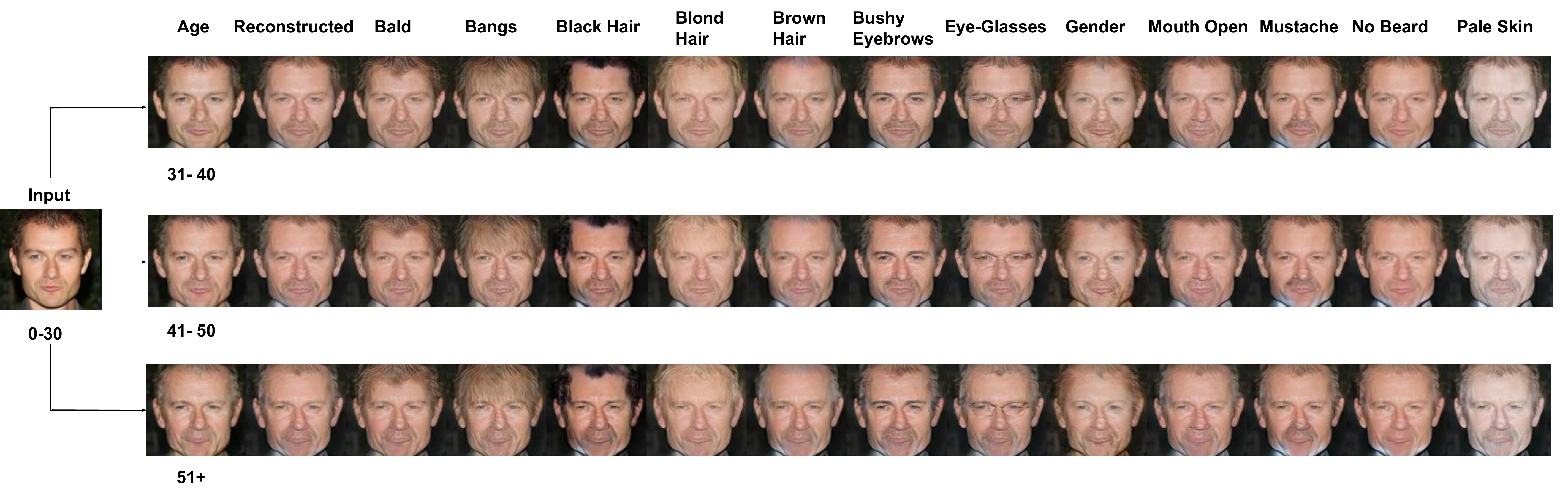}
  \vspace{-0.8cm}
  \caption{On CelebA dataset.}
\end{subfigure}
\begin{subfigure}{\linewidth}
\centering
  \includegraphics[width=\linewidth]{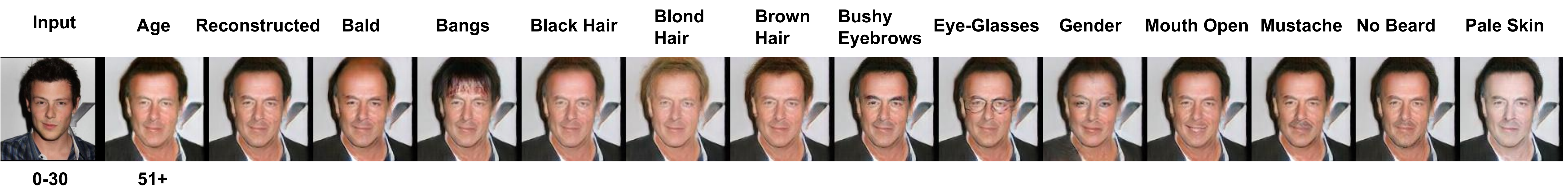}
  \vspace{-0.8cm}
  \caption{On CACD dataset.}
\end{subfigure}
\vspace{-0.2cm}
\caption{Results of our proposed FAWAM model. (a) On CelebA dataset: For a given input, the first image in each output row shows the aged face and the remaining images are attribute manipulated for respective age.(b) On CACD dataset: Attribute manipulated images corresponding to the [51+] age cluster.}
\label{Fig:our_result}
\end{figure*}

\subsubsection{\textbf{Quantitative results:}}For evaluating the facial attribute editing accuracy, we use facial attribute classifier trained on CelebA dataset for the generated faces. If the generated image attribute is same as the desired one by the classifier then it is a valid generation. Fig. \ref{fig:fam_acc} shows the attribute accuracy, as can be seen the model achieves good results for bangs, gender, pale skin, eyeglasses with 99.01\%, 91.6\%, 99.1\% and 88.3\% of accuracies, respectively. We can observe that bald, mustache, are less accurate due to the difficulty in modelling and less samples in dataset.

\subsection{FAWAM results:}
Here we explain the results of our proposed FAWAM model. For our problem at hand, we have two inputs viz, young image as well as the attribute. From the proposed model we first obtain the aged face from FA module and then apply attribute editing using AM module.
\vspace{-0.3cm}
\subsubsection{\textbf{Visualization results:}}
\textbf{Visualization results for our FAWAM model:}
FAWAM results for a single input image that produce multi-aged and single-attribute images as outputs are shown in Fig. \ref{Fig:our_result}. From the results, we can see that for different aging clusters, there is a subtle difference in their attributes produced owing to the aging factor, which is a significant realistic advantage of our proposed model.

\begin{figure}[bt]
\small 
\captionsetup{size=small}

\begin{subfigure}{\columnwidth}
\centering
  \includegraphics[width=1\linewidth]{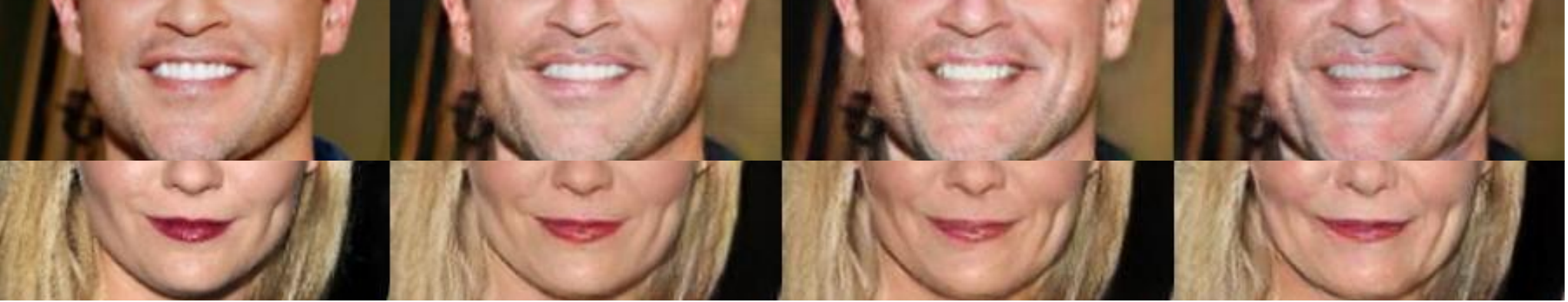}
  \caption{Aging Consistency }
\label{fig:d3}
\end{subfigure}
\begin{subfigure}{0.2\textwidth}
\centering
  \includegraphics[width=1\linewidth]{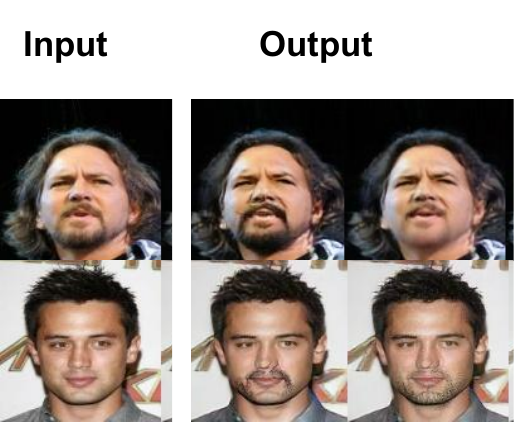}
  \caption{Attribute consistency for mustache and beard }
\label{fig:d1}
\end{subfigure}
\hspace{1cm}
\begin{subfigure}{0.2\textwidth}
\centering
  \includegraphics[width=1\linewidth]{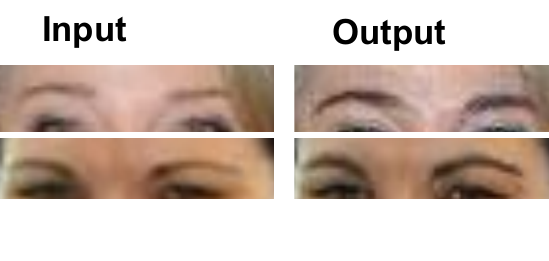}
  \caption{Attribute consistency for bushy eyebrows }
\label{fig:d2}
\end{subfigure}
\vspace{-0.2cm}
\caption{Illustration of visual fidelity}
\end{figure}

\noindent\textbf{Visual fidelity:}
Fig. \ref{fig:d3} displays face images showing cheeks region. We can observe that the synthesised images are photo-realistic that clearly depict the age-related features such as wrinkles, thin lips etc. Fig. \ref{fig:d1} and Fig. \ref{fig:d2} show the attributes incorporated along with aging for beard, mustache and bushy eyebrows respectively.

\vspace{-0.3cm}
\subsubsection{\textbf{Quantitative results:}}
\begin{figure}[bt]
\small 
\captionsetup{size=small}

\begin{center}
   \includegraphics[width=0.9\linewidth]{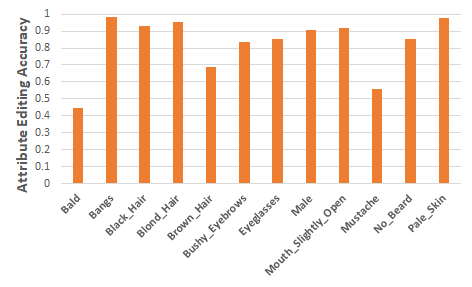}
\end{center}
\vspace{-0.8cm}
   \caption{Attribute editing accuracy of our proposed FAWAM model (Higher the Better).}
\label{fig:our_acc}
\end{figure}

\noindent\textbf{Accuracy of our FAWAM faces:}
Fig. \ref{fig:our_acc} shows the attribute accuracy of our FAWAM model, as can be seen the model achieves good results for bangs, gender, pale skin, eyeglasses with 98.07\%, 90.4\%, 97.6\% and 85.0\% of accuracies, respectively. We can observe that bald, mustache, brown hair are less accurate compared to others. We also perform the facial verification as shown in Table. \ref{table:our_Verf} on CelebA dataset for the test images. We perform verification only on CelebA dataset due the availability of attributes. Since this study marks one of the first works on \textit{face aging along with attribute manipulation}, the results in this paper bestow the \textit{state-of-the-art} for the problem.\\

\begin{figure}[h!]
\small 
\captionsetup{size=small}
\begin{center}
   \includegraphics[width=1\linewidth]{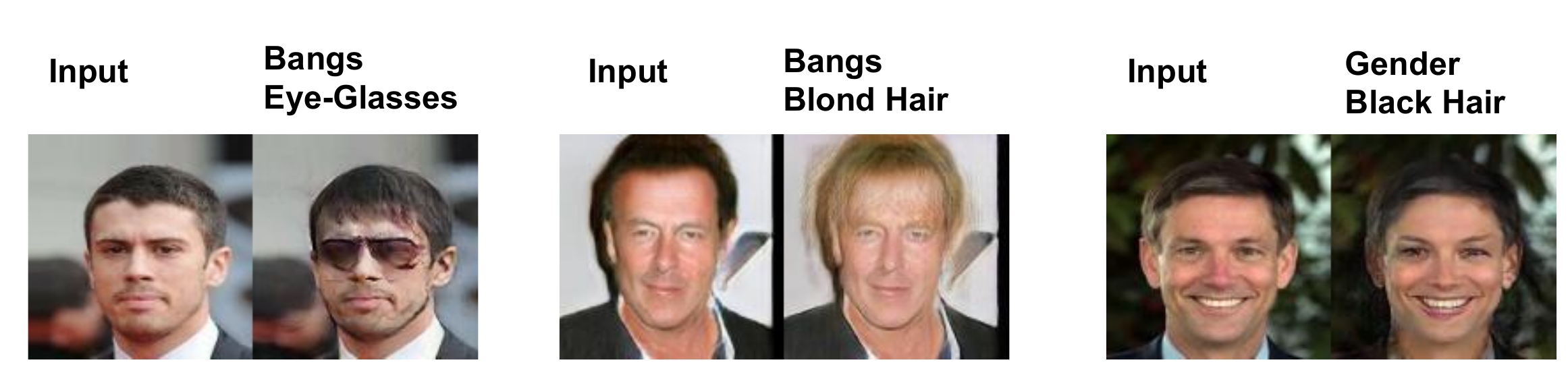}
\end{center}
\vspace{-0.5cm}
   \caption{Multi facial attribute editing on CelebA dataset.}
\label{fig:multi-att}
\end{figure}

\begin{figure}[bt]
\small 
\captionsetup{size=small}
\begin{center}
   \includegraphics[width=1\linewidth]{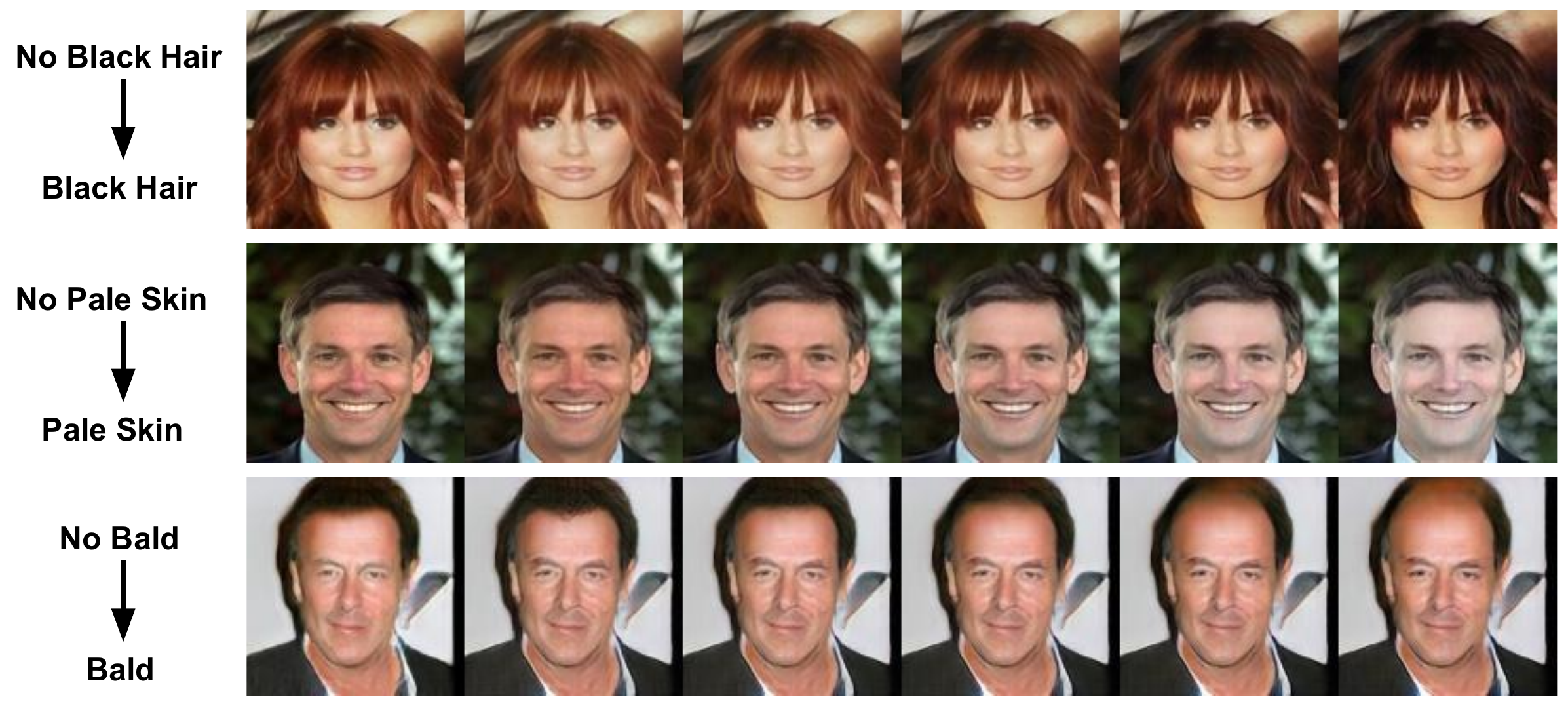}
\end{center}
\vspace{-0.5cm}
   \caption{Illustration of attribute intensity control.}
\label{fig:intensity}
\end{figure}

\vspace{-0.2cm}
\noindent\textbf{Ablation studies:}
Here we demonstrate various ablation studies carried out on our model. Fisrt, we show the possibiity of \textbf{multi-attribute editing}, i.e. to manipulate multiple facial attributes simultaneously. The resulting images are shown in Fig. \ref{fig:multi-att}. Another ablation study is conducted to see the \textbf{impact of attribute intensity}. In this regard, we perform attribute intensity control. Although the attributes are binary, we generalise to continuous attribute value while testing. The results for the same are shown in Fig. \ref{fig:intensity}, for an attribute intensity control of 20\% per image across the row. We can see that the simulated output results are smooth and continuous, highlighting the effectiveness of our model.

\begin{table}[bp]
    \centering
\small 
\captionsetup{size=small}
\caption{Face verification results for our proposed FAWAM model (values are in \%)}
\label{table:our_Verf}
    \begin{tabular}{c l c c c}
    \toprule
    \multirow{3}{*}{\bfseries Dataset} & 
    \multirow{3}{*}{\bfseries Test} & 
    \multicolumn{3}{c}{\bfseries Verification Confidence}\\ \cmidrule(lr){3-5}
    && Age[31-40] & Age[41-50] & Age[51+] \\ 
    \cmidrule(lr){1-5}
         &  Young   & 92.5 & 88.6 & 84.02 \\
    CelebA &  Old  & - & 91.2 & 89.4 \\
    \bottomrule
\end{tabular}
\end{table}

\section{Conclusions}
In this paper, we proposed a novel holistic model towards Face age progression with attribute manipulation (FAWAM). In particular, we tackle the task by leveraging two submodules viz., face aging (FA) module and attribute manipulation (AM) module, which are independent of each other and provide equal importance while image rendering. For the FA module, leveraging pyramidal GAN, the model is able to come up with age-specific changes along with identity preservation. Regarding the AM module, the synthesised aged face further undergoes attribute manipulation with the aid of attribute classification constraint, adversarial learning and reconstruction learning for desired change of attributes, visually realistic editing and to preserve attribute-excluding details respectively. Extensive analysis of the model is carried out in a bottom up manner and the achieved attribute-manipulated aged imageries and the quantitative evaluations clearly confirm the effectiveness and robustness of the proposed method.
Our future works include dual-discriminator based GAN for further improvements.

{\small
\bibliographystyle{ieee_fullname}
\bibliography{egbib}
}

\end{document}